\documentclass[runningheads]{llncs}
\usepackage{graphicx}
\usepackage[export]{adjustbox}
\usepackage{amsmath}
\usepackage{amssymb}
\usepackage{bbm}
\usepackage{tabularx}
\usepackage{makecell} 
\usepackage[ruled,vlined]{algorithm2e}
\usepackage{nameref}
\usepackage{enumitem}
\usepackage{multirow}

\begin{document}
%
\title{Multi-task Graph Convolutional Neural Network for Calcification Morphology and Distribution Analysis in Mammograms}


\author{Hao Du\inst{1}
\and
Melissa Min-Szu Yao\inst{2,3}
\and 
Liangyu Chen\inst{5}
\and 
Wing P. Chan\inst{2,3,4}
\and
Mengling Feng\inst{1,6}}
\authorrunning{Du et al.}

\institute{Saw Swee Hock School of Public Health, National University of Singapore, Singapore 117549 \\
\email{duhao@u.nus.edu}
\and
Department of Radiology, School of Medicine, College of Medicine, Taipei Medical University, Taiwan
\and
Department of Radiology, Wan Fang Hospital, Taipei Medical University, Taipei 116, Taiwan
\and
Medical Innovation Development Center, Wan Fang Hospital, Taipei Medical University, Taiwan
\and
School of Electrical and Electronic Engineering, Nanyang Technological University, Singapore
\and
Institute of Data Science, National University of Singapore \\
\email{ephfm@nus.edu.sg}
}

\maketitle
%

\begin{abstract}

%
The morphology and distribution of microcalcifications in a cluster are the most important characteristics for radiologists to diagnose breast cancer. However, it is time-consuming and difficult for radiologists to identify these characteristics, and there also lacks of effective solutions for automatic characterization. In this study, we proposed a multi-task deep graph convolutional network (GCN) method for the automatic characterization of morphology and distribution of microcalcifications in mammograms. Our proposed method transforms morphology and distribution characterization into node and graph classification problem and learns the representations concurrently. Through extensive experiments, we demonstrate significant improvements with the proposed multi-task GCN comparing to the baselines. Moreover, the achieved improvements can be related to and enhance clinical understandings. We explore, for the first time, the application of GCNs in microcalcification characterization that suggests the potential of graph learning for more robust understanding of medical images. 

\end{abstract}

\section{Introduction}

According to Global Cancer Statistics 2020, breast cancer has overtaken lung cancer as the most common cancer around world \cite{sung2021global}. Nevertheless, the good news is that the 5-year survival rate for breast cancer can be as high as 90\% if it is detected early before it progress to metastatic cancer \cite{cokkinides2005american}. Mammography is currently the most effective tool for early detection of breast cancer, and it is widely adopted in breast cancer screening \cite{misra2010screening}. In mammographic based breast cancer diagnosis, microcalcification (MC) clusters are an important early sign that accounts for approximately 50\% of  the diagnosed cases \cite{scimeca2014microcalcifications,baker2010new}. An MC cluster contains at least 3 individual MCs where each MC is a small amount of calcium deposits in breast tissue and appears as a small bright spot in mammograms \cite{ma2010novel}. Mammography images commonly have high resolution, which enables the detection of MCs at an early stage \cite{chen2014topological}. 

\begin{figure}[!t]
\caption{Examples of morphology and distribution types. Types of suspicious morphology include coarse heterogeneous, fine pleomorphic, amorphous and fine linear (fine-linear branching). The types of distribution  includes diffuse, regional, cluster(grouped), linear and segmental.}
\centering
\includegraphics[width=0.9\textwidth, frame]{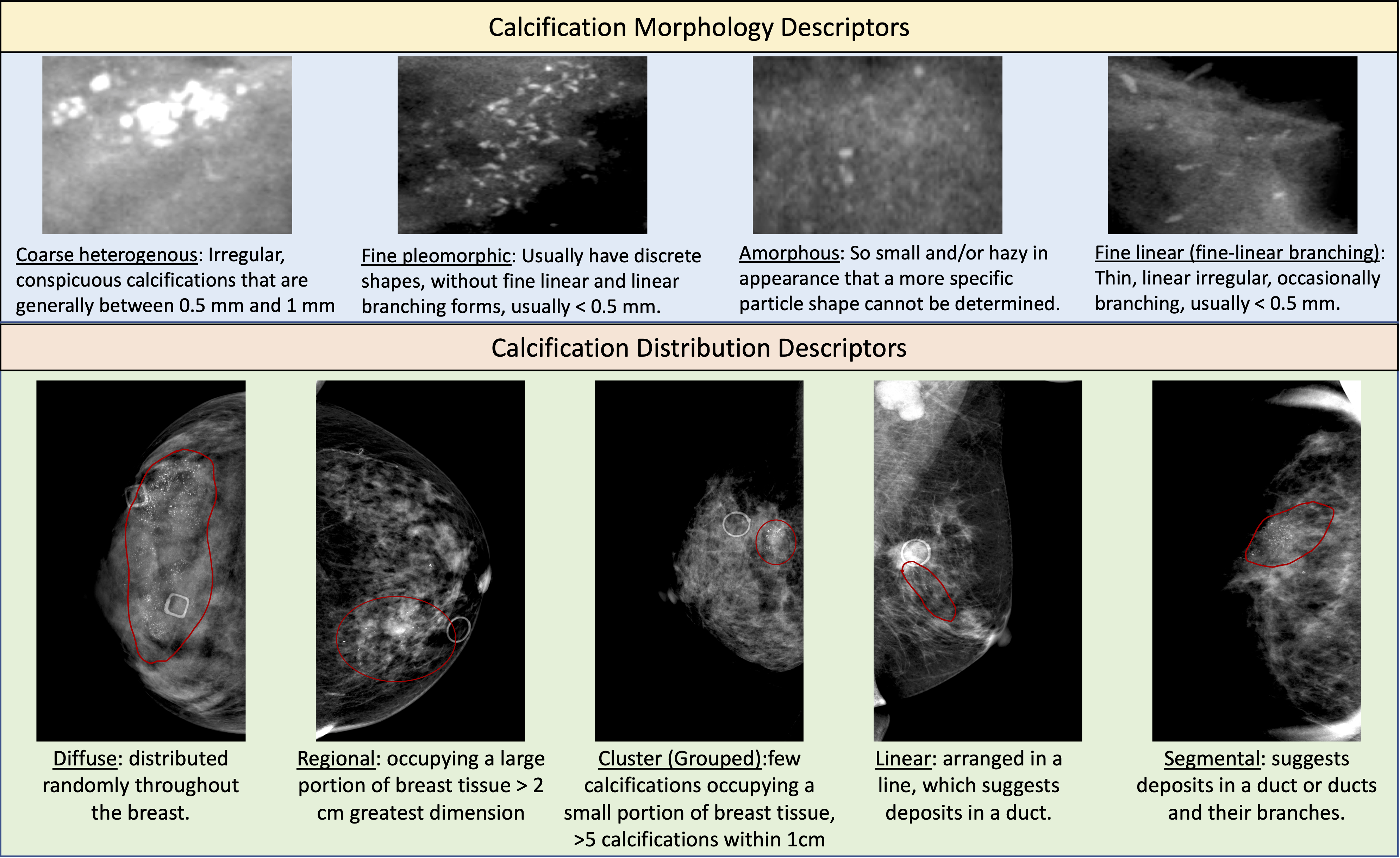}
\label{fig:figure1}
\end{figure}

However, only certain types of MCs are associated with a high probability of malignancy \cite{dahnert2017radiology}. The American College of Radiology Breast Imaging Reporting and Data System (ACR BI-RADS) classifies calcifications into two categories: typically benign or suspicious, according to the morphology and distribution of calcifications \cite{american2013acr}. Morphology describes the form of calcifications based on shape, size, brightness, roughness etc. Distribution describes how calcifications spread throughout breast tissue. The morphology and distribution of calcifications, illustrated in Figure \ref{fig:figure1}, are the most important characteristics considered by radiologist to provide appropriate follow-up recommendations.

Recently, numerous computer-aided diagnosis (CADx) methods have been developed to classify calcifications into benign or malignant clusters \cite{alam2019classification,singh2018approach,alam2018automatic,papadopoulos2005characterization,chen2014topological,oliver2012automatic,ciecholewski2017microcalcification,strange2014modelling,shao2011characterizing}. 
Alam et al. \cite{alam2019classification,alam2018automatic} selected calcification density, distances from cluster centroids, cluster areas and calcification sizes to discriminate between benign and malignant calcification clusters. Singh et al. \cite{singh2018approach} utilized shape and texture features to determine malignancy. Although the effectiveness of these features has been proven, existing CADx methods are unable to characterize the MCs into the descriptors of morphology and distribution, as recommended by ACR BI-RADS \cite{american2013acr}. Automatic characterization of calcifications is important to reproduce the chain of reasoning for mammogram interpretation, leading to more accurate and robust understanding of mammograms.

To address this challenge, we formulate the characterization of calcifications in mammograms as a multi-task classification problem and propose a graph convolutional neural network (GCN) framework. Firstly, we transform the calcifications in mammography images to graphical data to represent the spatial and visual information. That is, each calcification is represented by a node and nodes are connected according to the geometric relationships. Following the transformation, we formulate the morphology classification as a node classification task and distribution classification as a graph classification task. We propose a multi-task model with GCNs to solve both tasks. By employing GCNs \cite{defferrard2016convolutional,kipf2016semi,li2019deepgcns}, we incorporate both local patch features and topological structure. 
Multi-task learning provides shared representation between different tasks, improves the proposed model's generalizability. We summarize our contributions as follows: 
%
\begin{enumerate}
  \item We transform information of calcifications in mammography images into graphical representations. 
  \item We propose a deep GCN based framework to model the node and graph embeddings for both morphology and distribution tasks.
  \item We develop a GCN-based solution to characterize both morphology and distribution with multi-task training strategy. With extensive experiments, we showed that the proposed multi-task training strategy leads to better performance compared to models trained on a single task and other baseline models.

\end{enumerate}
%
%
%
%
\section{Methodology}

\begin{figure}[h]
\caption{Proposed framework demonstration. (a) Illustration of graph construction for calcification clusters. (b) Illustration of multi-task deep GCN with inputs from (a). }
\centering
\includegraphics[width=\textwidth, frame]{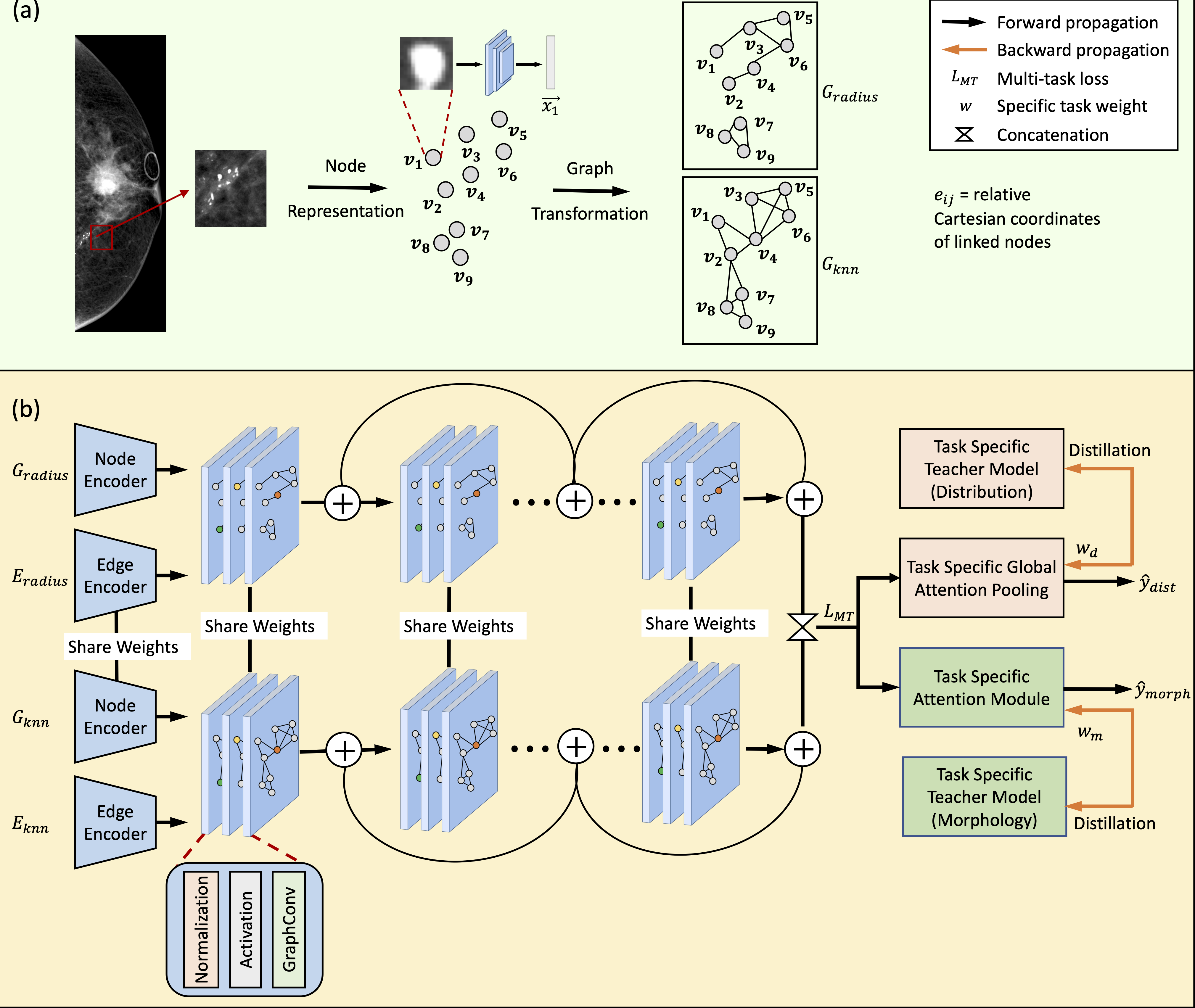}
\label{fig:figure2}
\end{figure}

The structure of proposed model is divided into graph construction and multi-task GCN, as illustrated in Figure \ref{fig:figure2}. In the first step, we transform the calcifications in mammography images into graphical data by using convolutional neural network (CNN) as feature extractor and graph transformation functions. Following graph construction, the proposed GCN jointly learns representations for node and graph classification with multi-task training strategy.

Let $x^{I}$ be a mammography image, $x^{c}$ be the set of calcifications in the image. A set of $N$ mammography images  $X = \{x_i\}^{N}_{i=1}$ where $x_{i} = (x_{i}^{I}, x_{i}^{c})$ are included in our dataset. We transform image set $X$ to graphical set $\mathcal{G}$ with $G_{i} \in \mathcal{G}$ and $G = (\mathcal{V}, \mathcal{E})$, where $\mathcal{V} = \{v_1, v_2, \dots, v_N \}$ and $\mathcal{E} \subseteq \mathcal{V} \times \mathcal{V}$ are the sets of vertices and edges, respectively. $e_{ij}$ represents an edge connecting vertices $v_i$ and $v_j$ if the edge $e_{ij} \in \mathcal{E}$. A vertex $v$ and an edge $e$ in the graph are associated with vertex features $h_{v} \in \mathbb{R}^{D}$ and edge features $h_e \in \mathbb{R}^{C}$ respectively, where $D$ and $C$ are dimensions of vertex and edge features. There are two tasks to investigate: (1) node (morphology) classification, where each vertex $v$ has a label $y_v$ and we aim to learn function $f$ and representation $r_v$ such that the vertex label could be predicted as $y_v = f(r_v)$; (2) graph (distribution) classification, where the graph has a label $y_{g}$ and we aim to learn function $g$ and representation vector $r_g$ to predict the label of the graph as $y_g = g(r_g)$.

\subsection{Graph construction}
\label{gc}

Graph construction is demonstrated in part (a) of Figure \ref{fig:figure2}. For each mammography image with calcifications $(x_{i}^{I}, x_{i}^{c})$, we define a set of patches as $P = \{p_1, p_2, \dots, p_n\}$, where $p$ represents an image patch that locates at the center of a calcification with dimension $M \times M$. We extract high level features from patches $P$ with a convolutional neural network (CNN) as a feature extractor. We concatenate extracted features with the normalized coordinates of the patches to form the node feature $h_{v}$. The edge features $h_{e}$ are defined as relative Cartesian coordinates of linked nodes. Following node and edge feature extraction, we construct two types of graphs based on the spatial connectivity relationship between calcifications: 

\begin{enumerate}
	\item K-nearest neighbor (KNN) graph $G_{knn}$: Creates edges if the nodes are within the $k$  nearest neighbors. KNN graphs have been widely adopted in point cloud classification and segmentation  \cite{wang2019dynamic,landrieu2018large,te2018rgcnn}, image classification \cite{monti2017geometric}, etc. However, it may cause information loss from disconnected neighbors in dense calcification clusters or introduce noise when the node is an outlier from the calcification cluster. 

	\item Radius graph $G_{radius}$: Creates edges based on node positions to all other nodes within a given distance. The radius graph solves the limitations introduced by the KNN graph described above, but it is affected by a constant distance threshold which may cause information loss for vertices beyond the threshold. 

\end{enumerate}








\subsection{Deep Graph Convolutional Network}

As illustrated in Figure \ref{fig:figure2}~(b), we stack GCN blocks to construct deep GCN. Following \cite{li2019deepgcns} and \cite{li2020deepergcn}, we use GCN blocks with Normalization $\rightarrow$ ReLU $\rightarrow$ GraphConv $\rightarrow$ Addition and GENeralized Aggregation Networks (GENconv) as GraphConv backbone. In GENconv, the message construction function $p^{(l)}$ is defined to apply on vertex feature $h_v^{(l)}$, neighbor vertex's feature $h_u^{(l)}$ and edge feature $h_{e_{vu}}^{(l)}$ to construct the message to propagate. $p^{(l)}$ is defined as:

\begin{equation} 
\label{eq:1}
m_{vu}^{(l)} = \rho^{(l)} (h_v^{(l)}, h_u^{(l)}, h_{e_{vu}}^{(l)}) = \text{ReLU} (h_u^{(l)} + \mathbbm{1}(h_{e_{vu}}^{(l)}) \cdot h_{e_{vu}}^{(l)}) + \epsilon, u \in \mathcal{N}(v)
\end{equation}
where the ReLU$(\cdot)$ represents the rectified linear unit activation function \cite{nair2010rectified}, $\mathbbm{1}(\cdot)$ is an indicator function which equals to 1 when edge features exist otherwise 0, and $\epsilon$ is a small positive constant. $\text{SoftMaxAgg}_{\beta}$ is then used as the message aggregation function and defined as: 
\begin{equation}
\label{eq:2}
m_v^{(l)} = SoftMaxAgg_{\beta}(\cdot) = \sum_{u \in \mathcal{N}(v)} \frac{ exp(\beta m_{vu}^{(l)}) }{ \sum_{i \in \mathcal{N}(v)} exp(\beta m_{vu}^{(l)}) } \cdot m_{vu}^{(l)},
\end{equation}

\noindent where $\mathcal{N}(v)$ is the set of neighbors of vertex $v$ and $\beta$ is a hyper-parameter which controls the aggregation function. Message normalization \textit{MsgNorm} is then introduced to address the oversmoothing and gradient vanishing problem in training deep GCNs. MsgNorm normalizes the features of the aggregated message $m_v^{(l)}$ by combining them with other features during the vertex update phase. Suppose MsgNorm is applied to a multi-layer perceptron (MLP) vertex update function $\text{MLP}(h_v^{(l)} + m_v^{(l)})$, the vertex update function becomes as follows:
\begin{equation}
\label{eq:3}
h_v^{l+1} = \phi^{l(l)} ( h_v^{(l)}, m_v^{(l)} ) = \text{MLP}( h_v^{(l)} + s \cdot \lVert h_v^{(l)} \rVert_2 \cdot \frac{m_v^{(l)}}{\lVert m_v^{(l)} \rVert_2 } )
\end{equation}
where $s$ is a learnable scaling factor. The aggregated message $m_v^{(l)}$ is first normalized by its $\ell_2$ norm and then scaled by the $\ell_2$ norm of $h_v^{(l)}$ by a factor of $s$. The scaling factor $s$ is set to be a learnable scalar with an initialized value of 1.

\subsection{Multi-task Learning}

In this study, the proposed multi-task GCN is trained to jointly perform morphology and distribution classification. In general, the model is trained by a multi-task loss $L_{MT} = w_{m}L_{m} + w_{d}L_{d} $ where $w_mL_{m}$ and $w_dL_{d} $ are weighted cross-entropy loss for morphology and distribution classification, respectively. In the ACR BI-RADS guideline, morphology and distribution of calcifications are equally important. Therefore, we introduced GradNorm \cite{chen2018gradnorm} to learn both tasks at an equal pace. Firstly, we define the necessary quantities as below: 


\begin{itemize}
  \item $W$: The subset of the full network weights $W \subset \mathcal{W}$. The weights of the last shared layer is generally chosen as $W$. 
  \item $G_W^{(i)}(t) = \Vert \nabla_{W} w_i(t) L_i(t) \Vert_{2}$: the $L_2$ norm of the gradient over the weighted loss $w_i(t) L_i(t)$  for task $i$ with respect to $W$, at training step $t$.

  \item $\overline{G}_{W}(t) = E_{task}[G_{i}^{W}(t)]$: the average value of gradient norms over all tasks for training step $t$. 
  \item $\tilde{L}_i(t) = \frac{L_i(t)}{L_i(0)}  $: the loss ratio as the inverse training rate of task $i$ at step $t$;
  \item $r_i(t) = \frac{\tilde{L}_i(t)}{E_{task}[\tilde{L}_i(t)] }$: the relative inverse training rate of task $i$ at step $t$.
\end{itemize}

In order to balance the gradient magnitudes $G_W^{(i)}$ for each task, the mean gradient norm across all tasks $\overline{G}_{W}$ is set as the common scale target. The relative inverse training rate of task $i$, $r_i(t)$, is used to balance the learning pace of all tasks. The target gradient norm for task $i$ is: 
\begin{equation}
\label{eq:5}
	G_W^{(i)}(t) \rightarrow \overline{G}_W(t) \times [r_i(t)]^{\alpha},
\end{equation}
\noindent where $\alpha$ controls the strength of the restoring force which pulls tasks back to a common training rate. A higher value of $\alpha$ indicates a higher strength to enforce training rates to be balanced. 

Equation \ref{eq:5} provides the target gradient norms for task $i$. At each training step $t$, we update the loss weights $w_i(t)$ to bring gradient norms close to the target for task $i$. $L_1$ loss between the actual gradient norms and the target at each time step for each task is introduced as $L_{grad}$ and we sum $L_{grad}$ across both morphology and distribution classification tasks. 

\begin{equation}
\label{eq:6}
L_{grad}(t; w_i(t)) = \sum_{i} \vert G_{W}^{(i)}(t) - \overline{G}_{W}(t) \times [r_i(t)]^{\alpha} \vert_1
\end{equation}

Similar to \cite{liu2019improving}, we apply knowledge distillation \cite{hinton2015distilling} method to further improve the multi-task network. We train multiple models  with outstanding performance in morphology and distribution tasks as teacher models. We then train a single multi-task network to distillate knowledge with the cross-entropy loss on the soft targets generated from these teacher models. 
%
\section{Experiments and Results}

\textbf{Dataset} The full field digital mammogram dataset for this study was collected from \textit{Anonymous Organization}. The dataset contains 387 mammography images from 200 patients who are classified as ACR BI-RADS category 4 and 5 with documented calcifications from the original radiological reports. All cases were confirmed breast cancers from biopsy tests. Descriptors of morphology and distribution were annotated by a senior radiological technologist and carefully reviewed by two senior radiologists in a joint meeting. The dataset was divided into 80\% as training set and 20\% as testing set. The codes are available at xxxx. 

\begin{table*}
	\begin{center}
		\begin{tabularx}{\textwidth}{X | X | X | X} 
			\hline
			\textbf{Type} & \textbf{Methods} & \makecell[l]{\textbf{Distribution} \\ \textbf{AUC}} & \makecell[l]{\textbf{Morphology} \\ \textbf{AUC}} \\
			\hline\hline
			\multirow{5}{*}{Baseline} & ResNet \cite{he2016deep} &  0.683 & 0.565 \\
			\cline{2-4}
			& DenseNet \cite{huang2017densely} & 0.688 & 0.576  \\ 
			\cline{2-4}
			& MobileNet \cite{howard2017mobilenets} & 0.694 & 0.575 \\ 
			\cline{2-4}
			& ShuffleNet \cite{zhang2018shufflenet} & 0.678 & 0.581 \\ 
			\hline

			\multirow{6}{*}{Ablation Study} & \makecell[l]{Task-specific (Mor.)} & ---  & 0.608 \\
			\cline{2-4}
			& \makecell[l]{Task-specific (Dis.) } & 0.785 & ---  \\ 
			\cline{2-4}
			& \makecell[l]{Single-graph (Rad.) } & 0.785 & 0.585 \\ 
			\cline{2-4}
			& \makecell[l]{Single-graph (KNN)} & 0.742 & 0.580 \\ 
			\cline{2-4}
			& \makecell[l]{2-layer GCN} & 0.769 & 0.597 \\ 
			\cline{2-4}
			& \makecell[l]{4-layer GCN} & 0.751 & 0.619 \\ 
			\cline{2-4}
			& \makecell[l]{16-layer GCN} & 0.802 & 0.615 \\ 
			\hline
			\textbf{Proposed}& \makecell[l]{\textbf{Multi-task,} \\ \textbf{multi-graph,} 
			\\
			\textbf{8-layer GCN}} & \textbf{0.815} & \textbf{0.631}\\
			\hline
		\end{tabularx}
	\end{center}
	\caption{The performance comparison between baseline models, ablation study models and proposed model on distribution and morphology classification. Mor.=Morphology, Dis.=Distribution, Rad.=Radius}
	\label{tab:resultall}
\end{table*}


\noindent \textbf{Performance Comparison} In our experiments, we used the multi-class AUC for performance evaluation \cite{fawcett2006introduction}. AUC was evaluated at the node and graph level for morphology and distribution classification, respectively. To the best of our knowledge, there is no state-of-the-art models to characterize morphology and distribution of calcifications into descriptors in mammography images. For baseline comparison, we regarded distribution classification as a multi-class classification problem over mammography images, and morphology classification as a multi-class classification problem over calcification patches, and employed multiple commonly used models in computer vision, shown in Table \ref{tab:resultall}:Baseline. Distribution baseline models took mammography images $X^{I}$ as input to predict the types of distribution. For morphology baseline models, we took mammography images $X^{I}$ and the set of patches $P$ defined in Section \ref{gc}. Similar to vertices in constructed calcification graphs, each patch was associated with a morphology label. The baseline models classified the patch set into morphology categories.

As Table \ref{tab:resultall} shows, our proposed model demonstrated leading performance across both tasks. The improvements on distribution classification task can be attributed to the design of GCN which captures the geometrical relationships between calcifications, thereby improving the ability to distinguish distribution types. For morphology, the improvements can be attributed to the message propagation from neighboring vertices with the same morphology type. Calcifications with the same morphology tend to locate in a nearby region or cluster. Therefore, the feature propagation from neighbors enhances the proposed model to distinguish morphology.

\noindent \textbf{Ablation experiments for multi-task network}. We separately trained task-specific models by removing the distribution or morphology branch respectively, as shown in Table \ref{tab:resultall}:Ablation Study. The multi-task model outperformed the task-specific models in both tasks. The improvements can be attributed to the fact that distribution and morphology are associated and jointly affect the radiologists' decision-making on malignancy diagnosis. For example, in ductal carcinoma in situ and invasive ductal carcinoma, fine linear or linear branching calcifications often have a segmental ductal distribution \cite{chen2012segmental}. Fine pleomorphic and linear branching calcifications in a segmental distribution are highly suspicious for malignancy \cite{chen2012segmental}.  The design of the multi-task network learns the shared representation from the morphology and distributed labels, thus achieving improvements on both tasks.

\noindent \textbf{Ablation experiments for depth of deep GCNs}. To investigate the effectiveness of depths of Deep GCN, we compared different number of graph convolutional layers in the proposed network in Table \ref{tab:resultall}:Ablation Study. The experiment results showed that relative larger number of GCN layers improves the performance. In GCNs, single layer of GCN considers nearest neighbor while networks with multiple GCN layers perform message propagation and fusion from multi-hop neighbors. As mentioned, calcifications with same morphology locate in a nearby region or cluster and distribution considers how calcifications spread over the breast. To a certain extent, when the depth of GCN increases, message propagation from more hops of neighbors enhances the network's ability in classifying nodes and graphs. However, when the network depth increases further, the message propagation from further nodes may be harmful for morphology classification because the further nodes may not have the same type of morphology. Deeper GCN in this study may also suffer from oversmoothing and gradient vanishing problems, which could be investigated in future studies. 


\noindent \textbf{Ablation experiments for multi-graph fusion}. To investigate the effectiveness of multi-graph fusion, we compared with multi-task model with single radius or KNN graph as input to GCN in Table \ref{tab:resultall}:Ablation Study. The experiment results showed that the multi-graph fusion improves the robustness of the model. As mentioned in Section \ref{gc}, individual graph has limitations in either morphology or distribution classification task. The design of multi-graph fusion enhances the model's ability to learn representations from two graphs, thereby improving on both classification tasks. 

\section{Conclusions}

We proposed a multi-task GCN model to tackle the challenging problem of characterization of calcifications in mammography images. Characterizing the distribution and morphology is essential to apply computerized assisted detection tools in mammography. 

\clearpage
\bibliographystyle{splncs04}
\bibliography{mtcalsbib.bib}
\end{document}